\pdfoutput=1

\documentclass[11pt,twocolumn]{article}

\usepackage[preprint]{acl}

\usepackage{times}
\usepackage{latexsym}

\usepackage[T1]{fontenc}

\usepackage[utf8]{inputenc}

\usepackage{microtype}

\usepackage{inconsolata}

\usepackage{graphics}
\usepackage[utf8]{inputenc} 
\usepackage[T1]{fontenc}    
\usepackage{url}            
\usepackage{booktabs}       
\usepackage{amsfonts}       
\usepackage{nicefrac}       
\usepackage{microtype}      
\usepackage[utf8]{inputenc}
\usepackage{graphicx}
\usepackage{amsmath}
\usepackage{amssymb}
\usepackage{mathtools}
\usepackage{amsthm}
\usepackage{arydshln}
\usepackage{multirow}
\usepackage{wrapfig, lipsum, booktabs}
\usepackage{algorithm}
\usepackage{enumitem}
\usepackage{paralist, tabularx}
\usepackage{balance}
\usepackage[noend]{algpseudocode}
\usepackage{pgfplots}
\usetikzlibrary{pgfplots.groupplots}
\pgfplotsset{compat=1.3}
\usepackage{tikz}
\usetikzlibrary{patterns}
\usepackage{pgf-pie}
\usepackage{adjustbox}
\usepackage{colortbl}
\usepackage{xspace}
\usepackage{titling}
\PassOptionsToPackage{table}{xcolor}
\usepackage[most]{tcolorbox}

\newcommand{\impro}[1]{{\hspace{0.05cm}{\color[HTML]{32CB00}\textbf{(+#1)}}}}

\definecolor{demphcolor}{RGB}{144, 144, 144}
\definecolor{mygray}{gray}{0.4}
\definecolor{lightgray}{rgb}{0.9, 0.9, 0.9}

\definecolor{dt}{HTML}{ADCAD8}
\definecolor{dt2}{HTML}{cddfe7}

\usepackage{pifont}
\usepackage{float}
\definecolor{my_green}{RGB}{51,102,0}
\definecolor{my_red}{RGB}{204, 0, 0}

\DeclareMathAlphabet{\mathsfit}{\encodingdefault}{\sfdefault}{m}{sl}
\SetMathAlphabet{\mathsfit}{bold}{\encodingdefault}{\sfdefault}{bx}{n}

\definecolor{oorange}{RGB}{252,218,227}
\definecolor{yyellow}{RGB}{255,237,203}
\definecolor{ppurple}{RGB}{208,205,226}
\definecolor{ggreen}{RGB}{195,222,176}
\definecolor{ggrey}{RGB}{230,230,230}
\definecolor{rred}{RGB}{247,187,187}
\definecolor{wwhite}{RGB}{255,255,255}

\newcommand{\modelname}{RICO\xspace}
\newcommand{\modeldpo}{RICO-Flash\xspace}

\title{\modelname: Improving Accuracy and Completeness in Image Recaptioning via Visual Reconstruction}

\author{Yuchi Wang$^{1}$, ~Yishuo Cai$^{2}$, ~Shuhuai Ren$^{1}$, ~Sihan Yang$^{3}$, ~Linli Yao$^1$ ,~Yuanxin Liu$^1$, \\ \textbf{Yuanxing Zhang$^4$,~Pengfei Wan$^4$,~Xu Sun$^1$} \\
 $^{1}$ National Key Laboratory for Multimedia Information Processing, Peking University \\
 $^{2}$Central South University 
   $^{3}$Xi'an JiaoTong University 
   $~^{4}$Kuaishou Technology \\
   \texttt{wangyuchi@stu.pku.edu.cn
   ~xusun@pku.edu.cn} \\
   ~\\
}

\makeatletter
\AtBeginDocument{
\let\@oldmaketitle\@maketitle
\renewcommand{\@maketitle}{\@oldmaketitle
  \vspace{-28pt}
  \includegraphics[width=1\linewidth]{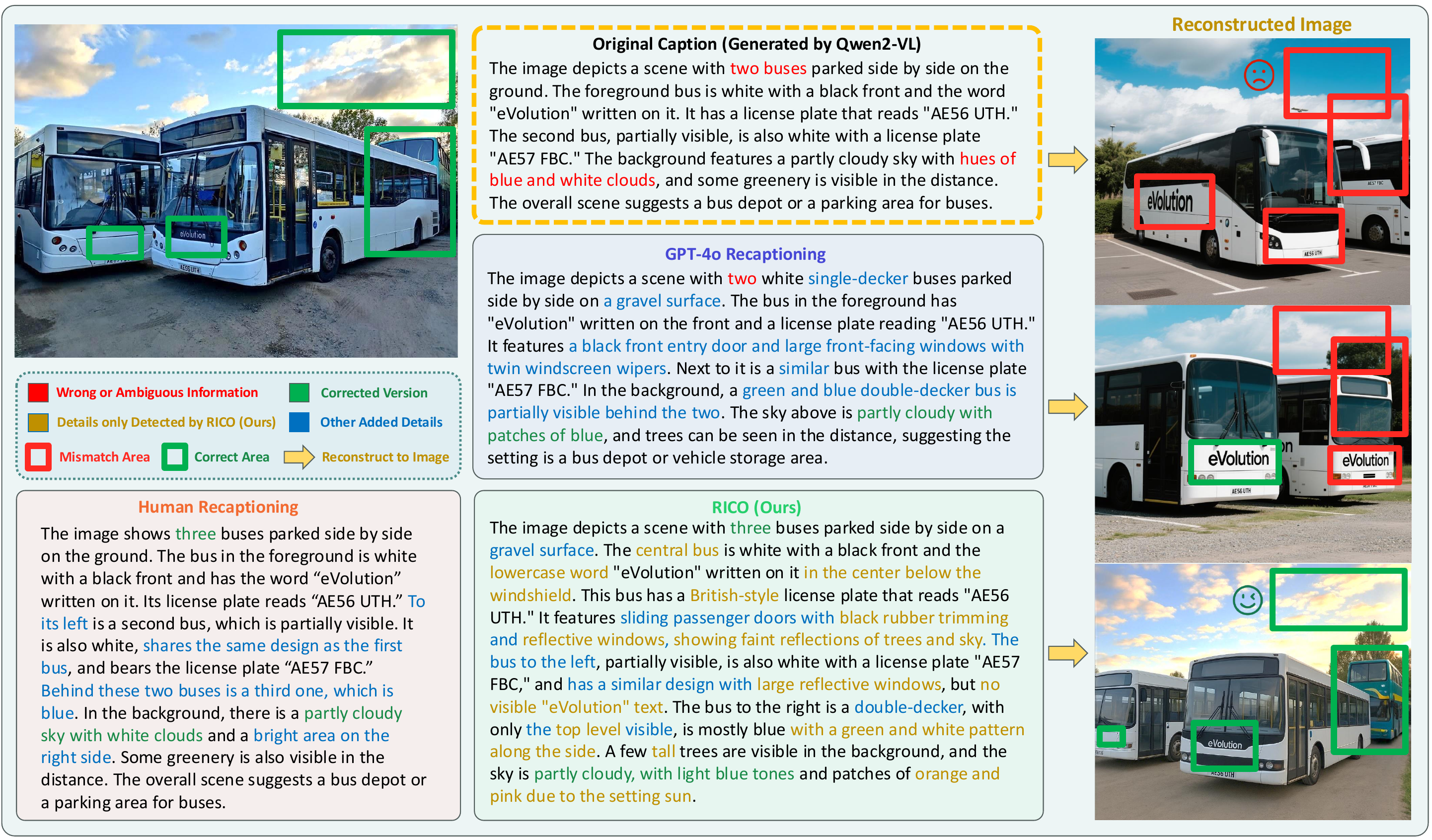}
  \vspace{-18pt}
  \captionof{figure}{
  Analysis of image captions generated by Qwen2-VL and its recaptioned variants. Despite the advanced capabilities of Qwen2-VL, the generated captions still contain incorrect or ambiguous information---for example, misidentifying the number of buses---a mistake that remains uncorrected even by GPT-4o. Furthermore, both GPT-4o and human-generated recaptions often overlook fine-grained details, such as attributes and spatial relationships, which are accurately captured by our model. By reconstructing images from captions, it becomes evident that our model better preserves such details, resulting in reconstructions that more closely resemble the original image.
}
  \label{fig:intro}
  \vspace{13pt}
 }
}
\makeatother

\begin{document}

\maketitle

\begin{abstract}
Image recaptioning is widely used to generate training datasets with enhanced quality for various multimodal tasks. Existing recaptioning methods typically rely on powerful multimodal large language models (MLLMs) to enhance textual descriptions, but often suffer from inaccuracies due to hallucinations and incompleteness caused by missing fine-grained details. 
To address these limitations, we propose \modelname, a novel framework that refines captions through visual reconstruction. Specifically, we leverage a text-to-image model to reconstruct a caption into a reference image, and prompt an MLLM to identify discrepancies between the original and reconstructed images to refine the caption. This process is performed iteratively, further progressively promoting the generation of more faithful and comprehensive descriptions.
To mitigate the additional computational cost induced by the iterative process, we introduce RICO-Flash, which learns to generate captions like RICO using DPO. Extensive experiments demonstrate that our approach significantly improves caption accuracy and completeness, outperforms most baselines by approximately 10\% on both CapsBench and CompreCap. Code released at \url{https://github.com/wangyuchi369/RICO}.

\end{abstract}

\section{Introduction} \label{sec:intro}

The availability of hundreds of millions of image-text pairs collected from the internet has played a pivotal role in advancing modern multimodal learning~\cite{sharegpt4v,liu2023visualinstructiontuning,bai2023qwenvlversatilevisionlanguagemodel}. However, the alt text associated with web images is frequently of low quality, offering uninformative descriptions or even text unrelated to the image content. 
Consequently, recaptioning methods have been widely employed to generate enhanced captions for downstream multimodal tasks, such as training multimodal large language models (MLLMs)~\cite{sharegpt4v}, text-to-image models~\cite{dalle-3}, and CLIP-like models~\cite{laclip,veclip}.

Typically, recaptioning methods primarily depend on powerful MLLMs~\cite{veclip, sharegpt4v}. While MLLMs significantly enhance captions over the alt-text by leveraging their strong perceptual capabilities, the generated descriptions still face two key challenges: (1) \textbf{Inaccuracy}, where some descriptions are incorrect, often exacerbated by the notorious hallucination problem of MLLMs~\cite{bai2025hallucinationmultimodallargelanguage}; and (2) \textbf{Incompleteness}, where important details are frequently omitted. 
These issues cannot be fully resolved even with the integration of additional models or human editing. For example, as illustrated in Fig.~\ref{fig:intro}, the caption generated by Qwen2-VL~\cite{qwen2vl} contains ambiguous or incorrect information that cannot be fully corrected even with GPT-4o~\cite{openai2024gpt4ocard}. Moreover, several visual details remain undetected by either GPT-4o or human annotators, whereas our method successfully captures them. This appears to stem from the natural tendency of both humans and models to focus on salient objects in an image, often neglecting attributes and subtle details. 
We further validate this observation through experiments in \textsection~\ref{sec:exp_eff}.

\begin{figure*}
    \centering    \includegraphics[width=0.95\linewidth]{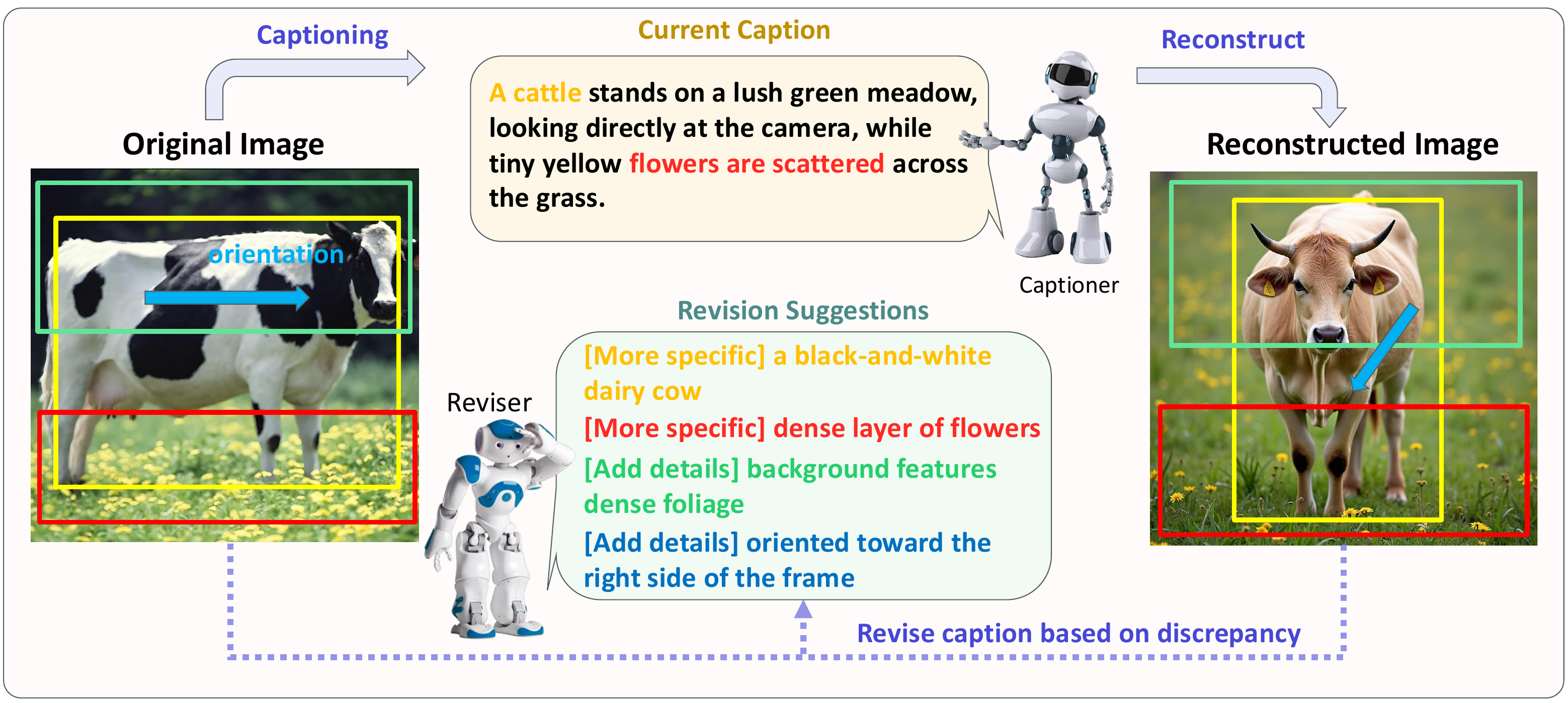}
    \caption{Illustration of the motivation for introducing the visual reconstruction mechanism. Conventional recaptioning methods typically map images directly to text without explicitly aligning the semantic spaces of the two modalities, often leading to information loss in the generated captions. In contrast, our approach incorporates visual reconstruction to make this loss more observable. By identifying discrepancies between the original and reconstructed images through the reviser, we refine the caption to produce a more semantically aligned and comprehensive description.
}
    \label{fig:t2i}
\end{figure*}

From a semantic space perspective, the challenges above suggest that the semantic space constructed through recaptioning is often biased and lossy compared to that of the original image. As illustrated in Fig.~\ref{fig:t2i}, conventional captioners typically follow a one-way mapping from image to text, without enforcing explicit semantic alignment between the two modalities, resulting in the omission of critical semantic elements in the generated captions.
We argue that an ideal cross-modal semantic alignment should involve a bi-directional mapping: when text is generated from an image, the reconstructed image from that text should remain consistent with the original. In cases of misalignment, the discrepancy between the original and reconstructed images can be used to adjust the semantic space of the caption. Based on this intuition, we propose \modelname (\textbf{R}econstruction-guided \textbf{I}mage \textbf{C}aption \textbf{O}ptimization), a novel recaptioning framework. As shown in Fig.~\ref{fig:t2i}, our method incorporates a visual reconstruction step that makes semantic discrepancies more observable in the visual domain compared to simply contrasting image and text, thereby facilitating the recovery of omitted details and producing descriptions that are both more semantically aligned and comprehensive.

Technically, we use powerful text-to-image models to reconstruct each caption into a reference image. Next, we input the original image, the generated reference image, and the candidate caption into a reviser---an MLLM and prompt it to refine the caption based on the discrepancies between the original and reference images. Through experiments, we find that a single-step refinement is insufficient, so we design the refinement process to iterate multiple times to progressively improve the caption.
Given the significant time and computational resources required for iterative refinement, we propose an end-to-end variant as a more efficient alternative to \modelname. This model is constructed by learning the naturally induced preference relationships during the iterative refinement process using Direct Preference Optimization (DPO)~\cite{rafailov2024direct}. Specifically, we employ \modelname to generate a batch of training data, which is then used to fine-tune a base model via DPO, resulting in the compact \modeldpo model.

Through experiments, we demonstrate that our pipeline effectively constructs well-aligned image-text information spaces. From the captioning perspective, we evaluate both the \modelname framework and the compact \modeldpo model on some benchmarks. Results show that \modelname significantly enhances caption quality in terms of both accuracy and comprehensiveness. For instance, it consistently achieves improvements of over 10 points on CapsBench~\cite{capsbench}. Moreover, \modeldpo outperforms all recaptioning baselines.
From the reverse perspective of text-to-image generation, we find that models trained on captions refined by \modeldpo exhibit a stronger understanding of fine-grained prompts, particularly with regard to attributes and relationships. Further analysis also reveals that our method demonstrates strong robustness and generalization across diverse settings.


\section{Related Works}

\subsection{Multimodal Large Language Models}
Inspired by the success of large language models (LLMs)~\cite{sun2025descriptivecaptionenhancementvisual,ouyang2022traininglanguagemodelsfollow,deepseekai2025deepseekr1incentivizingreasoningcapability,bai2023qwentechnicalreport} in natural language processing, several works have extended them to multimodal settings by incorporating visual encoders~\cite{gpt4v,liu2023visualinstructiontuning,geminiteam2024geminifamilyhighlycapable,bai2023qwenvlversatilevisionlanguagemodel,ren2024timechattimesensitivemultimodallarge}, contributing to multimodal large language models (MLLMs). Flamingo~\cite{alayrac2022flamingovisuallanguagemodel} is an early effort that inserts gated attention layers into a pretrained language model to enable vision-language understanding. Subsequent works explore various strategies for connecting vision encoders to language models. For example, BLIP-2~\cite{li2023blip2bootstrappinglanguageimagepretraining} introduces the Q-Former to bridge the modalities, LLaVA~\cite{liu2023visualinstructiontuning} employs a simple MLP projection layer, and Gemini~\cite{geminiteam2024geminifamilyhighlycapable} feeds image and text tokens jointly into a unified Transformer.
In addition to architectural design, recent research has also focused on improving the quality of pretraining and fine-tuning data~\cite{bai2023qwenvlversatilevisionlanguagemodel,wang2024cogvlmvisualexpertpretrained,zhu2023minigpt4enhancingvisionlanguageunderstanding}. While modern MLLMs demonstrate impressive visual perception capabilities, they still suffer from hallucination issues~\cite{bai2025hallucinationmultimodallargelanguage}---occasionally generating inaccurate or fabricated content---which undermines the faithfulness of the generated captions.

\subsection{Image Recaptioning}
Describing an image using text has been a fundamental task in multimodal learning~\cite{li2022blipbootstrappinglanguageimagepretraining,Ghandi_2023,wang2024ladicdiffusionmodelsreally,yao2023capenrichenrichingcaptionsemantics}. Among these efforts, image recaptioning aims to generate enhanced captions for original, noisy alt text associated with image-text pairs. It has become increasingly important for producing high-quality synthetic data to support various downstream applications. This trend was popularized by DALL-E 3~\cite{dalle-3}, which introduced the idea of replacing low-quality or overly simplistic captions with synthetic alternatives. Since then, numerous approaches have leveraged image recaptioning to improve multimodal large language models (MLLMs)~\cite{sharegpt4v}, text-to-image generation models~\cite{dalle-3}, and CLIP-style vision-language models~\cite{veclip, laclip}.
Among these efforts, LaCLIP~\cite{laclip} utilizes LLMs to rewrite alt-text, while VeCLIP~\cite{veclip} incorporates additional visual information. CapsFusion~\cite{capsfusion} trains a LLaMA-based model to fuse alt-text and synthetic captions, and ShareGPT4V~\cite{sharegpt4v} directly generates new captions using GPT-4V~\cite{gpt4v}. More sophisticated approaches include Altogether~\cite{xu2024altogetherimagecaptioningrealigning}, which employs iterative human annotation, and~\citeauthor{ye2025paintingwordselevatingdetailed} propose automated fine-grained feedback mechanisms to improve captioning capabilities. Additionally, methods based on local perception have also been explored~\cite{peng2025patchmatterstrainingfreefinegrained,sun2025descriptivecaptionenhancementvisual}.
However, despite their advancements, these methods fundamentally follow a paradigm of directly generating captions without explicitly enforcing semantic alignment between visual and textual modalities, inevitably resulting in considerable information loss.

\section{Methodology}

In this section, we introduce our \modelname framework and \modeldpo model. \textsection~\ref{sec:method_over} provides an overview of the pipeline of \modelname. Subsequently, \textsection~\ref{sec:method_init} describes how we generate the reference reconstruction image. \textsection~\ref{sec:method_con} presents the method designed to refine the caption. Finally, \textsection~\ref{sec:method_dpo} illustrates the process of training a compact model \modeldpo to learn the iterative process using DPO.

\subsection{Overall Pipeline of \modelname} \label{sec:method_over}

\begin{figure}
    \centering
    \includegraphics[width=0.95\linewidth]{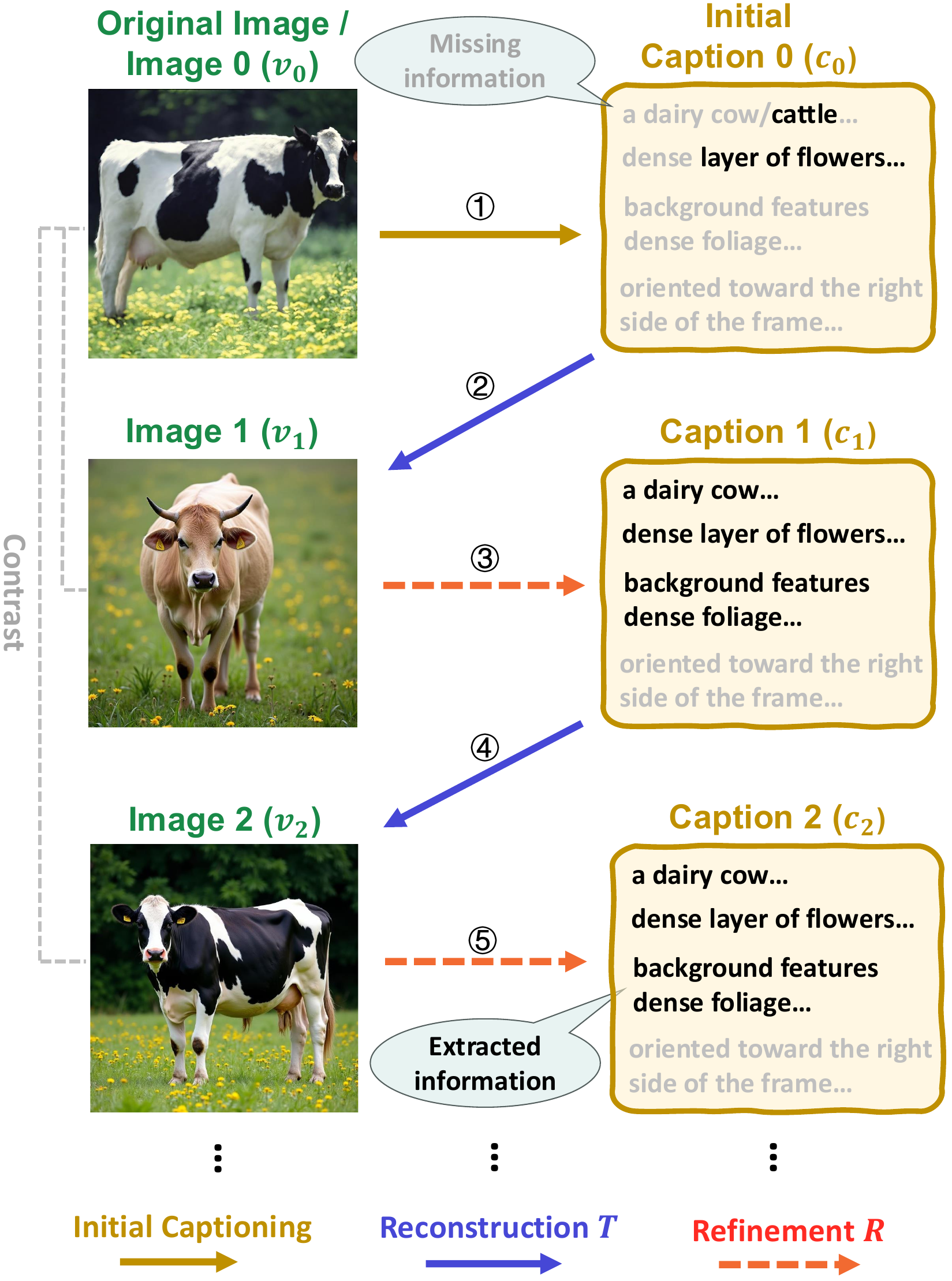}
    \caption{Illustration of the iterative process of \modelname. After the \textbf{\textcolor{brown}{initial captioning}} step, a \textbf{\textcolor{blue}{reconstruction}} procedure is applied to generate an image from the candidate caption. The caption is then \textbf{\textcolor{red}{refined}} by comparing the original image with the reconstructed image.}
    \label{fig:model}
\end{figure}

As illustrated in Fig.~\ref{fig:model}, in our \modelname framework, the initial caption \( c_0 \) for the original image \( v_0 \) is generated by the initial captioning model. A reconstruction model \( \mathbf{T} \) and a refinement model \( \mathbf{R} \) are then alternately applied to iteratively improve the caption. In each iteration $i \geq 1$, the reconstruction procedure converts the previous candidate caption \( c_{i-1} \) into a reconstructed image \( v_i \), and the refinement model generates a refined caption based on the previous caption $c_{i-1}$, the original image $v_0$, and the reconstructed reference image $v_i$. Formally, the refinement step is defined as:
\[
c_{i} = \mathbf{R}(v_i, v_0, c_{i-1}) = \mathbf{R}(\mathbf{T}(c_{i-1}), v_0, c_{i-1}).
\]

\subsection{Reconstruct Candidate Caption into Reference Image} \label{sec:method_init}

As discussed in \textsection~\ref{sec:intro}, the semantic information space of captions generated by typical captioning processes tends to be biased and lossy compared to the information contained in the original image. Specifically, we denote the semantic space of the original image as $\mathcal{V}$ and that of the generated caption as $\mathcal{C}$. A biased caption implies that for some information $i \in \mathcal{C}$, $f(i) \notin \mathcal{V}$, and a lossy caption implies that for some information $j \in \mathcal{V}$, $g(j) \notin \mathcal{C}$, where $f$ represents the mapping from textual to visual information, and $g$ denotes the reverse.
A key insight of this work is that directly comparing the information spaces $\mathcal{V}$ and $\mathcal{C}$ is challenging due to the cross-modal nature of $f$ and $g$. To address this, we leverage a powerful text-to-image model to reconstruct the caption into an image. This enables a more direct comparison between the original image $\mathcal{V}$ and the reconstructed image $\hat{\mathcal{V}}$, as both reside in the visual modality. 

In particular, we use the FLUX.1-dev model~\cite{flux2024} as our text-to-image generator, given its strong performance and open-source availability. A notable advantage of FLUX.1-dev is its use of a T5 text encoder~\cite{raffel2023exploringlimitstransferlearning}, which supports longer prompts, surpassing the 77-token limit imposed by CLIP-based models. This allows us to process more detailed captions and faithfully reconstruct their visual content. Formally, for a given generated caption $c_{i-1}$, we use the text-to-image model to produce a reference image $v_i$ via $v_i = \mathbf{T}(c_{i-1})$, effectively translating the information space of the candidate caption into visual form and facilitating the identification of discrepancies from the original image.

\subsection{Refine Caption with Reference Image Feedback} \label{sec:method_con}

Having obtained the reconstructed reference image $v_i$, we proceed to refine the previous candidate caption $c_{i-1}$ based on the discrepancy between the reconstructed image $v_i$ and the original image $v_0$, thereby generating an updated caption $c_{i}$, defined as $c_{i} = \mathbf{R}(c_{i-1}, v_i, v_0)$. Given the complexity of this task, we utilize one of the most advanced multimodal large language models, GPT-4o~\cite{openai2024gpt4ocard}, to perform the refinement process.
We observed that directly feeding all relevant information into the model yields suboptimal results, highlighting the importance of prompt engineering. To address this, we carefully design prompts with attention to several key aspects outlined below. The complete prompt is provided in \textsection~\ref{sec:app_prompt}.

\noindent\textbf{Task Description~}
We explicitly inform the model of the task objective, with a particular emphasis on how the reference image is generated. Additionally, the model is instructed to focus on the discrepancies between the reference image and the original image as the basis for refining the caption.

\noindent\textbf{Aspects the Model Should Focus On~}
It is not intuitive for the refinement model to determine what aspects of the discrepancy between the original image and the generated reference image it should focus on, and ranking the importance of different aspects is challenging. Therefore, we provide the model with some guidance. We define eight aspects for the model to prioritize, including: \textit{`Visual Details, Composition \& Layout, Human Attributes (if applicable), Perspective \& Style, Text in the Image, Image Quality, World Knowledge, and Color Aesthetics.'}

\noindent\textbf{Guidance for Improvement Method~}
To guide the model in refining the candidate caption, we categorize improvements into two types: addressing \textit{inaccuracy} and \textit{incompleteness}. For inaccuracy, the model is instructed to identify and correct errors based on discrepancies between the original and reconstructed images, and to revise any ambiguous descriptions in the previous caption that may have caused inaccurate reconstruction. For incompleteness, the model is encouraged to incorporate missing details and to elaborate on key attributes of the main objects, such as color, shape, and other fine-grained characteristics.

\noindent\textbf{Force Model to Output Analysis Process~}
Inspired by the success of Chain of Thought (CoT)~\cite{wei2023chainofthoughtpromptingelicitsreasoning}, we prompt the model to output not only the revised caption but also the corresponding analysis process. This technique serves two purposes: it allows us to examine the reasoning steps of the black-box multimodal large language model, and, as shown in our experiments in \textsection~\ref{sec:abl}, it improves the quality of the generated captions by encouraging the model to deliberate more deeply. For practical implementation, we instruct the model to enclose the analysis within special markers \texttt{<analysis>}...\texttt{</analysis>} to facilitate automated post-processing.

\subsection{\modeldpo: Leverage DPO to Mitigate Computational Cost} 
\label{sec:method_dpo}

\noindent\textbf{Preliminaries of DPO~}
Direct Preference Optimization (DPO)~\cite{rafailov2024direct} is a recently proposed algorithm for aligning language models with human preferences without relying on reinforcement learning. Unlike traditional Reinforcement Learning from Human Feedback (RLHF) methods, which involve separate reward modeling and policy optimization steps, DPO formulates preference learning as a binary classification problem between preferred and dispreferred responses.
Formally, given a prompt $x$ and a pair of responses $(y^+, y^-)$, where $y^+$ is preferred over $y^-$, DPO optimizes the likelihood ratio between the two responses under a learned policy $\pi_\theta$ and a fixed reference policy $\pi_{\mathrm{ref}}$, using the following objective:
\begin{equation*}
\fontsize{9.5pt}{10pt}\selectfont
\begin{split}
\mathcal{L}_\text{DPO} &= 
-\mathbb{E}_{(x, y^+, y^-)\sim \mathcal{D}}\left[
\log \sigma \left(\beta \log \frac{\pi_{\theta}(y^+\mid x)}{\pi_\text{ref}(y^+\mid x)} \right.\right. \\
&\quad \left.\left. - \beta \log \frac{\pi_{\theta}(y^-\mid x)}{\pi_\text{ref}(y^-\mid x)}\right) \right],
\end{split}
\label{eq:DPO}
\end{equation*}
\normalsize
Here, $\beta$ is a temperature-like hyperparameter that controls the sharpness of the preference modeling. The objective encourages the model to assign higher relative likelihoods to preferred responses compared to dispreferred ones, with respect to the reference policy.

\begin{table*}[]
\small
\tabcolsep=0.17cm
\caption{Performance of \modelname and \modeldpo under different initial MLLM recaptioning models. For \modeldpo, we use the corresponding MLLM as the base model. In CapsBench, \textit{Acc.} denotes overall accuracy, and \textit{Rel.Pos.} indicates relative position accuracy. In CompreCap, \textit{Obj.}, \textit{Pix.}, \textit{Attr.}, and \textit{Rel.} represent object coverage, pixel coverage, attribute score, and relation score, respectively. \textit{Over.} in Amber refers to overall performance (see \textsection~\ref{sec:app_bench} for details). Green text indicates improvements. \modelname demonstrates significant gains over the original captions, while \modeldpo achieves performance close to that of \modelname.
}
\label{tab:iter}
\begin{tabular}{l|lll|l|llll}
\toprule[1.5pt]
\multirow{2}{*}{\textbf{Method}} & \multicolumn{3}{c|}{CapsBench} & Amber & \multicolumn{4}{c}{CompreCap}  \\ \cmidrule{2-9}
     & Acc. $\uparrow$ & Color $\uparrow$ & Rel. Pos. $\uparrow$ & Over. $\uparrow$ & Obj. $\uparrow$ & Pix. $\uparrow$ & Attr. $\uparrow$ & Rel. $\uparrow$ \\ \midrule

Qwen2-VL Init. & 42.0 & 48.1 & 32.4 & 59.7 & 69.82 & 60.02 & 2.66 & 2.81 \\
+ \modeldpo 
& 55.3$_\impro{13.3}$ & 66.7$_\impro{18.6}$ & 55.1$_\impro{22.7}$ & 60.6$_\impro{0.9}$ & 74.80$_\impro{4.98}$ & \textbf{63.35}$_\impro{3.33}$ & 2.84$_\impro{0.18}$ & \textbf{2.84}$_\impro{0.03}$ \\
+ \modelname($N=2$) 
& \textbf{59.0}$_\impro{17.0}$ & \textbf{67.1}$_\impro{19.0}$ & \textbf{59.5}$_\impro{27.1}$ & \textbf{62.2}$_\impro{2.5}$ & \textbf{75.04}$_\impro{5.22}$ & 63.04$_\impro{3.02}$ & \textbf{2.85}$_\impro{0.19}$ & 2.82$_\impro{0.01}$ \\ \midrule

LLaVA-1.5 Init. & 29.5 & 27.8 & 18.1 & 44.7 & 57.14 & 44.48 & 2.02 & 2.38 \\
+ \modeldpo 
& 46.2$_\impro{16.7}$ & 49.6$_\impro{21.8}$ & 38.7$_\impro{20.6}$ & 53.1$_\impro{8.4}$ &  66.68$_\impro{9.54}$ & 56.52$_\impro{12.04}$ & 2.53$_\impro{0.51}$ & 2.43$_\impro{0.05}$ \\
+ \modelname($N=2$) 
& \textbf{53.1}$_\impro{23.6}$ & \textbf{61.1}$_\impro{33.3}$ & \textbf{48.1}$_\impro{30.0}$ & \textbf{59.7}$_\impro{15.0}$ & \textbf{76.38}$_\impro{19.24}$ & \textbf{61.49}$_\impro{17.01}$ & \textbf{2.82}$_\impro{0.80}$ & \textbf{2.82}$_\impro{0.44}$ \\
\bottomrule[1.5pt]
\end{tabular}
\end{table*}

Given that our iterative refinement process incurs substantial inference time and computational overhead, we explore the development of an end-to-end variant. Noting that the iterative procedure implicitly induces a preference relationship between captions, we adopt Direct Preference Optimization (DPO) to learn these preferences. Specifically, we collect a high-quality image dataset and apply \modelname to generate refined captions. For each image $v^{(i)}$, we extract the initial caption $c^{(i)}_0$ and the final caption after $N$ refinement steps, $c^{(i)}_N$, forming a preference tuple $(v^{(i)}, c^{(i)}_0, c^{(i)}_N)$. Based on our empirical observation that $c^{(i)}_N$ consistently outperforms $c^{(i)}_0$ in most cases, we treat this pairwise preference as supervision for DPO training.
We adopt Qwen2-VL~\cite{qwen2vl} as the base model and fine-tune it using the DPO objective, yielding an end-to-end variant we denote as \modeldpo. This model directly generates improved captions without requiring iterative alternation between a text-to-image model and a caption refinement module, thereby significantly reducing inference cost while maintaining competitive performance.

\section{Experiments}

\subsection{Setup}
\subsubsection{Implementation Details}

For the implementation details, the text-to-image generation is performed using the FLUX.1-dev model~\cite{flux2024}, while the caption refinement process is conducted with GPT-4o (24-08-06)~\cite{openai2024gpt4ocard}. We set the number of interaction steps $N = 2$, based on empirical observations that this configuration achieves a good balance between performance and computational efficiency.
For the DPO experiments, we initialize with the Qwen2-VL model and set the preference scaling parameter $\beta = 0.1$. The model is fine-tuned for 3 epochs with a learning rate of $\eta = 1.0 \times 10^{-5}$. More implementation details can be found in \textsection~\ref{sec:app_details}.

\begin{table}[t] 
\small
\begin{center}
\tabcolsep=0.2cm
\caption{Recaptioning results by humans and models based on the initial caption. In our \modelname method, a single iteration of refinement is performed.
}
	\label{tab:human}
	\begin{tabular}{l|cccc}
		\toprule[1.5pt]
		 \multirow{2}{*}{\textbf{Model}} & \multicolumn{4}{c}{CapsBench (Subset)}   \\  
         \cmidrule{2-5}
        & Acc. & Color  & Rel. Pos. & Shape\\
        \midrule
        Original & 43.55 & 44.30 & 39.45 & 20.41 \\
        + GPT-4o Edit & 49.50 &53.02 &44.04 &24.49   \\
        + Human Edit & 50.96 & 51.30 & 47.82 & 27.01  \\
        + \modelname Edit & \textbf{54.08} & \textbf{65.47} & \textbf{34.04} & \textbf{49.51} \\
	\bottomrule[1.5pt]
	\end{tabular}

\end{center}

\end{table}

\subsubsection{Evaluation Benchmarks}
In the era of MLLMs, traditional captioning metrics~\cite{bleu, vedantam2015ciderconsensusbasedimagedescription} often fail to capture fine-grained details and inadequately penalize hallucinations. To address these limitations, in addition to the recently proposed reference-based metric CAPTURE~\cite{capture}, we adopt more advanced benchmarks to more faithfully evaluate the quality of our method. Specifically, we employ CapsBench~\cite{capsbench}, which uses QA pairs to assess the accuracy and comprehensiveness of generated captions. We also utilize CompreCap~\cite{comprecap}, which leverages a Directed Scene Graph to evaluate the correctness of object mentions and their relationships. Furthermore, we adopt Amber~\cite{amber} to assess hallucinations in the generated descriptions. More details can be found in \textsection~\ref{sec:app_bench}.

\subsection{Effectiveness of \modelname and \modeldpo} \label{sec:exp_eff}

\begin{table*}[]
\small
\caption{Comparison with baseline methods across various evaluation metrics. Our method achieves the best performance on most metrics, while \modeldpo demonstrates performance comparable to \modelname. Bold text indicates the best results, and underlined text denotes the second-best.
}
\label{tab:base}
\begin{tabular}{l|cccc|cccc|c|c}
\toprule[1.5pt]
\multirow{2}{*}{\textbf{Method}} & \multicolumn{4}{c|}{CapsBench}    & \multicolumn{4}{c|}{CompreCap}     &  \multicolumn{1}{c|}{Amber}    & \multirow{2}{*}{Capture}  \\ \cmidrule{2-10}
       
     & Acc. & Color & Shape & Rel. Pos. & Obj.    & Pix.   & Rel.    & Attr. & Over.  \\ \midrule
     
LaCLIP~(\citeauthor{laclip})       & 22.65  & 21.65    & 9.09  &  11.11 & 48.02   & 42.59    & 1.73   & 2.29    &    43.8     &      39.56          \\ 
CapsFusion~(\citeauthor{capsfusion})     & 35.04  & 38.14  &12.12   & 25.46  & 61.67 & 52.63      & 2.32   & 2.59   &  44.5   &     56.03           \\ 
Self-Loop~(\citeauthor{capture})     & 29.63  & 29.55  &  9.09   & 17.13   & 65.77    & 51.54  & 2.30    & 2.53    & 49.5           &     56.61      

\\ 

VeCLIP~(\citeauthor{veclip}) &25.19  & 27.84   &   11.11     & 13.43   & 49.60 &  42.25   &2.50  &     1.77       &     41.0    & 38.13 \\ 

ShareGPT4V~(\citeauthor{sharegpt4v}) &50.46  & 62.13   &   38.78   & 49.34  & 67.47 &  62.00   &2.83  &     2.81       &     56.2   & 59.80 \\ 
\midrule
    \modeldpo (Ours)    & \underline{55.32} & \underline{66.67} & \underline{50.29} & \underline{55.09} &    \underline{74.80}   & \textbf{63.35} & \underline{2.84} & \textbf{2.84}  & \underline{60.6} & \underline{65.52}   \\ 

\modelname (Ours)&    \textbf{59.02} & \textbf{67.14} & \textbf{53.68} & \textbf{59.51}  & \textbf{75.04} & \underline{63.04} & \textbf{2.85} & \underline{2.82} &        \textbf{62.2}    &  \textbf{65.98}            \\ \bottomrule[1.5pt]
\end{tabular}
\end{table*}

We verify that our \modelname{} pipeline effectively addresses both inaccuracy and incompleteness in recaptioning. Firstly, we use two popular open-source models Qwen2-VL~\cite{qwen2vl} and LLaVA-1.5~\cite{llava} as the initial captioning models to produce baseline captions, which are then refined by \modelname. As shown in Tab.~\ref{tab:iter}, even with just two refinement iterations, the captions generated by \modelname exhibit substantial improvements across all benchmarks and metrics. Notably, the improvement in the overall score on the Amber indicates that \modelname mitigates hallucination. Furthermore, on CapsBench, we emphasize two critical aspects---color and relative position---and show that the reconstruction step helps the model more accurately identify and correct fine-grained discrepancies.
In addition, we can see that \modeldpo achieves performance that closely matches \modelname while still demonstrating substantial improvements over the initial captions, validating its effectiveness as a non-iterative alternative.

Secondly, we assess recaptioning quality by comparing \modelname against GPT-4o and human annotators. We randomly select 100 images from CapsBench, generate initial captions using Qwen2-VL, and perform one round of editing using GPT-4o, \modelname, and human annotators. The results, shown in Tab.~\ref{tab:human}, demonstrate that \modelname achieves strong recaptioning performance, even surpassing humans, who tend to overlook fine-grained details. Some experiment details can be found in \textsection~\ref{sec:app_edit}.

Finally, we conduct a qualitative analysis of the refinement process and present examples showcasing the step-by-step improvement of captions through \modelname in \textsection~\ref{sec:app_qual}.

\subsection{Comparison with Other Recaptioning Methods}

We compare our approach with other recaptioning methods, and the results are presented in Tab.~\ref{tab:base}. \modelname demonstrates strong performance across all evaluation metrics, particularly in fine-grained aspects such as color, entity shape, and relative position. This highlights the importance of reconstruction for achieving better alignment between textual descriptions and visual content. Details on how the baseline methods perform recaptioning are provided in \textsection~\ref{sec:app_base}.

\subsection{Further Analysis}

We conduct more experiments to help better understand our \modelname pipeline.

\subsubsection{Verify Alignment via Text-to-Image Generation} \label{sec:exp_t2i}
\begin{table}[t] 
\vspace{-0mm}
\begin{center}
\small
\tabcolsep=0.10cm
\caption{Evaluation of a text-to-image generation model trained with original captions versus captions refined by our \modeldpo model. \textit{Rel.} and \textit{Attr.} represent relation and attribute respectively.
}
	\label{tab: t2i}
	\begin{tabular}{l|ccc|c}
		\toprule[1.5pt]
        \multirow{2}{*}{\textbf{Model}} & \multicolumn{3}{c|}{DPG-Bench}  &  \multirow{2}{*}{VQAScore}  \\   
         & Rel. & Attr. & Overall & \\
        \midrule

        FLUX w/ Init. Cap. 	& 89.95 &	80.08 & 78.50 & 0.84\\
        FLUX w/ \modeldpo 	& \textbf{90.55} &	\textbf{82.83}	& 	\textbf{80.34} & \textbf{0.85}\\

		\bottomrule[1.5pt]
	\end{tabular}
\end{center}

\end{table}

To verify that \modelname effectively builds a well-aligned image-text semantic space, we evaluate it on a classical downstream task: text-to-image generation.
We collect an image dataset from Huggingface\footnote{Mainly from \url{https://huggingface.co/datasets/jackyhate/text-to-image-2M}} and use \modelname to perform recaptioning. Specifically, for each image \( v \), we obtain both the initial caption \( c_0 \) and the refined caption \( c_N \), forming two datasets: \(\mathcal{D}_{\text{initial}} = \{(v^{(i)}, c_0^{(i)})\}\) and \(\mathcal{D}_{\text{refined}} = \{(v^{(i)}, c_N^{(i)})\}\). We then use these datasets to train two separate text-to-image generation models based on FLUX.1-dev. 
For evaluation, considering that the prompts in our dataset are typically long and thus incompatible with many existing benchmarks~\cite{ghosh2023genevalobjectfocusedframeworkevaluating, huang2025t2icompbenchenhancedcomprehensivebenchmark}, we adopt the recently proposed DPG-Bench~\cite{hu2024ellaequipdiffusionmodels}, which is designed to evaluate detailed prompts. Moreover, we also employ VQAScore~\cite{lin2024evaluatingtexttovisualgenerationimagetotext}---a reference-free metric that serves as a robust alternative to CLIPScore~\cite{hessel2022clipscorereferencefreeevaluationmetric,imagenteamgoogle2024imagen3}.
As shown in Tab.~\ref{tab: t2i}, the model fine-tuned on the refined dataset consistently outperforms the baseline across all metrics. Notably, it achieves improvements in entity, relation, and attribute dimensions, demonstrating that our reconstruction-refinement pipeline enhances the alignment between image and caption in fine-grained semantic aspects. Detailed training configurations are provided in \textsection~\ref{sec:app_t2i}.

\subsubsection{Saturation with Increased Iteration Steps}

\begin{figure}
    \centering
    \includegraphics[width=0.75\linewidth]{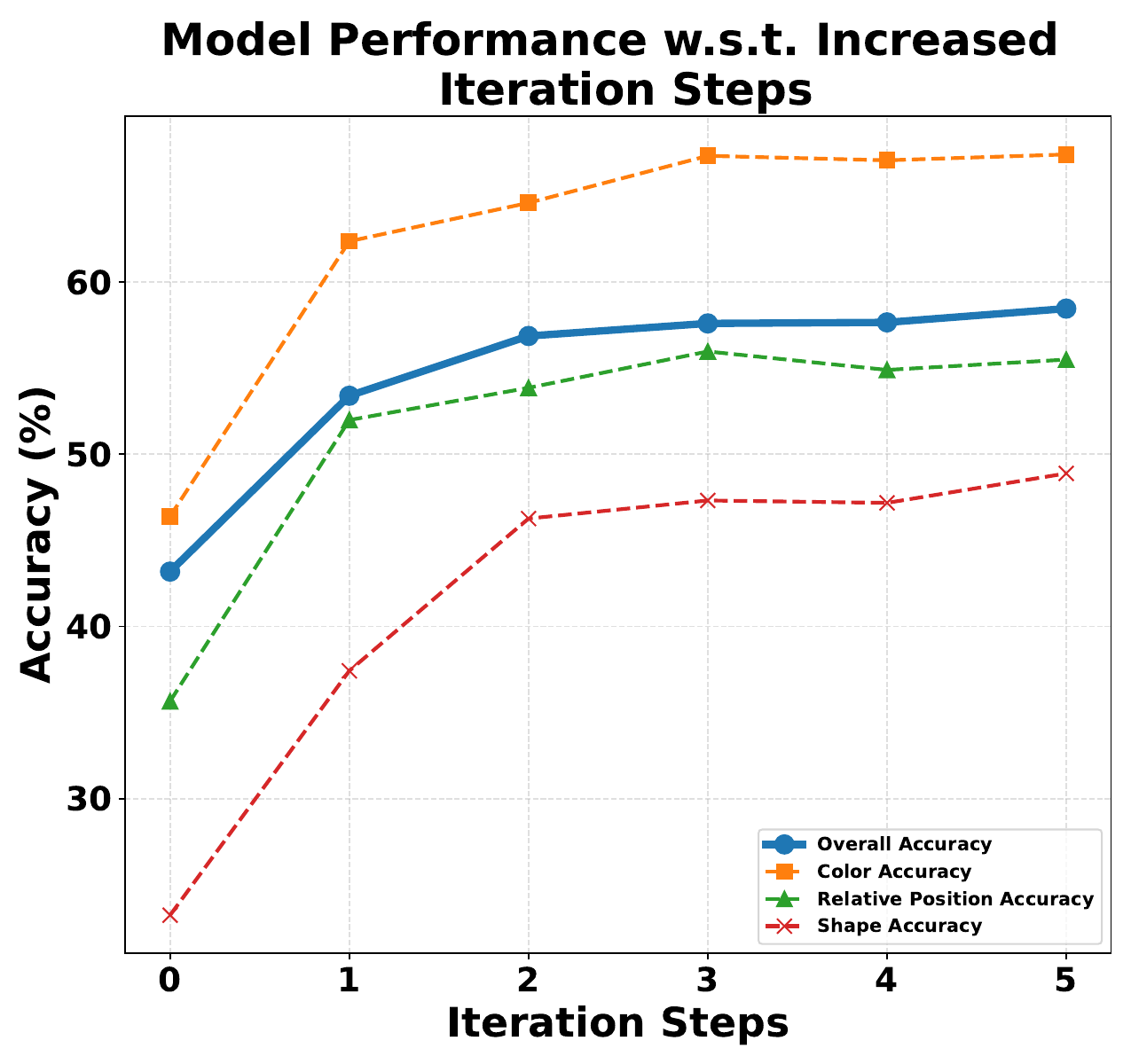}
    \caption{Performance of the \modelname pipeline under different numbers of refinement iterations.
}
    \label{fig:sat}
\end{figure}

In \modelname, the caption is progressively refined as the number of iteration steps increases. As shown in Fig.~\ref{fig:sat}, performance consistently improves with each additional iteration. However, the gains begin to plateau after approximately the second step, with only marginal improvements observed thereafter. This suggests that the generated caption reaches a satisfactory quality level, at least given the capabilities of the reconstruction and refinement modules.

\subsubsection{Generalization to Different Initial Captions} \label{sec:exp_general}

To evaluate the generalization capability of \modelname, we examine whether it can consistently enhance captions through the reconstruction-refinement pipeline. As shown in the upper part of Tab.~\ref{tab:rob}, we generate initial captions using different captioning models. The results indicate that \modelname significantly improves captions from all initial models, demonstrating its robustness. Notably, although our refinement module is based on GPT-4o, captions generated by GPT-4o alone do not outperform the final outputs, suggesting that \modelname does more than simply distill the captioning ability of GPT-4o.
In the lower part of Tab.~\ref{tab:rob}, we assess performance using various initial prompts. The results show that our pipeline yields substantial improvements across different prompts. While modifying the prompt within the same MLLM can lead to some gains, these are relatively minor compared to the improvements achieved by \modelname. Importantly, our method is also orthogonal to prompt-based strategies and can be combined with more effective prompts for further enhancement.

\begin{table}[t] 
\small
\begin{center}
\tabcolsep=0.08cm
\caption{Generality of \modelname across different initial recaptioning models and prompts. For CapsBench, we report overall accuracy, and for CompreCap, we use the unified metric for evaluation.}
\vspace{-1mm}
\label{tab:rob}
\begin{tabular}{l|c|c}
\toprule[1.5pt]
Model & CapsBench & CompreCap \\ 
\midrule
GPT-4o & 49.6 / 57.7$_\impro{8.1}$ & 58.6 / 60.4$_\impro{1.8}$ \\
Gemini 1.5 Pro & 49.7 / 57.7$_\impro{8.0}$ & 60.1 / 61.5$_\impro{1.4}$ \\
BLIP-3 & 37.0 / 56.2$_\impro{19.2}$ & 55.4 / 60.2$_\impro{4.8}$ \\
CogVLM 2 & 45.1 / 57.5$_\impro{12.4}$ & 56.0 / 60.3$_\impro{4.3}$ \\ 
\midrule
Qwen2-VL (Prompt 1) & 42.0 / 59.0$_\impro{17.0}$ & 55.9 / 61.4$_\impro{5.5}$ \\
Qwen2-VL (Prompt 2) & 46.0 / 57.6$_\impro{11.6}$ & 57.2 / 60.6$_\impro{3.4}$ \\
Qwen2-VL (Prompt 3) & 41.9 / 54.9$_\impro{13.0}$ & 56.9 / 60.9$_\impro{4.0}$ \\
\bottomrule[1.5pt]
\end{tabular}
\end{center}
\end{table}

\begin{table}[t] 
\small
\begin{center}
\tabcolsep=0.15cm
\caption{Ablation studies. 
}
\vspace{-1mm}
	\label{tab:model_abl}
		 \begin{tabular}{l|cccc}
		\toprule[1.5pt]
		 \multirow{2}{*}{\textbf{Method}} & \multicolumn{4}{c}{CapsBench}   \\  
         \cmidrule{2-5}
        & Acc. & Color  & Rel. Pos. & Shape\\
        \midrule 
        \modelname & \textbf{59.02}  & \textbf{67.14} & \textbf{59.51} & \textbf{53.68} \\
        \modeldpo  &\underline{55.32} & \underline{66.67} & \underline{55.09} & \underline{50.29}  \\
                \midrule
  (a) wo/ tips & 54.33 & 62.23 & 50.95 & 42.11\\
        (b) wo/ output analy. &50.40 & 62.54 & 53.24 & 36.36\\
        (c) finetune w/ pos.& 51.16 & 59.79 & 51.85 & 32.32\\
        (d) infer with ICL & 45.26 &49.83 &42.13 &26.26  \\

		\bottomrule[1.5pt]
	\end{tabular}

\end{center}

\end{table}
\subsection{Ablation Studies} \label{sec:abl}

We conduct ablation studies to validate our design choices, with results presented in Tab.~\ref{tab:model_abl}. The findings are: \textbf{(a)} When the refinement model is not guided on which aspects to focus, it struggles to identify key elements, resulting in a performance drop. \textbf{(b)} Omitting the requirement for the model to output an analysis process, which is intended to promote deliberate reasoning, also leads to degraded performance. 
Regarding the DPO method, we evaluate two alternative strategies: \textbf{(c)} directly fine-tuning the base model using positive samples, and \textbf{(d)} incorporating a positive sample into the prompt for in-context learning~\cite{dong2024surveyincontextlearning}. Both approaches yield inferior results compared to the DPO method, underscoring the effectiveness of DPO in our setting.

\section{Conclusion}
In this paper, we propose the \modelname pipeline, which leverages visual reconstruction to improve the accuracy and completeness of image recaptioning. We also introduce an efficient variant, \modeldpo, which learns the iterative refinement process of \modelname by DPO. Experimental results show that our method achieves well-aligned semantic representations between images and their captions, and delivers strong recaptioning performance compared to prior baselines.
Further evaluations also confirm the generalizability of our approach. We hope \modelname will inspire new techniques in image recaptioning and may contribute to advancements in broader multimodal research.

\bibliography{custom}

\newpage
\appendix

\section{Additional Experimental Results}
\subsection{Qualitative Analysis of \modelname} \label{sec:app_qual}

\begin{figure*}
    \centering    \includegraphics[width=1\linewidth]{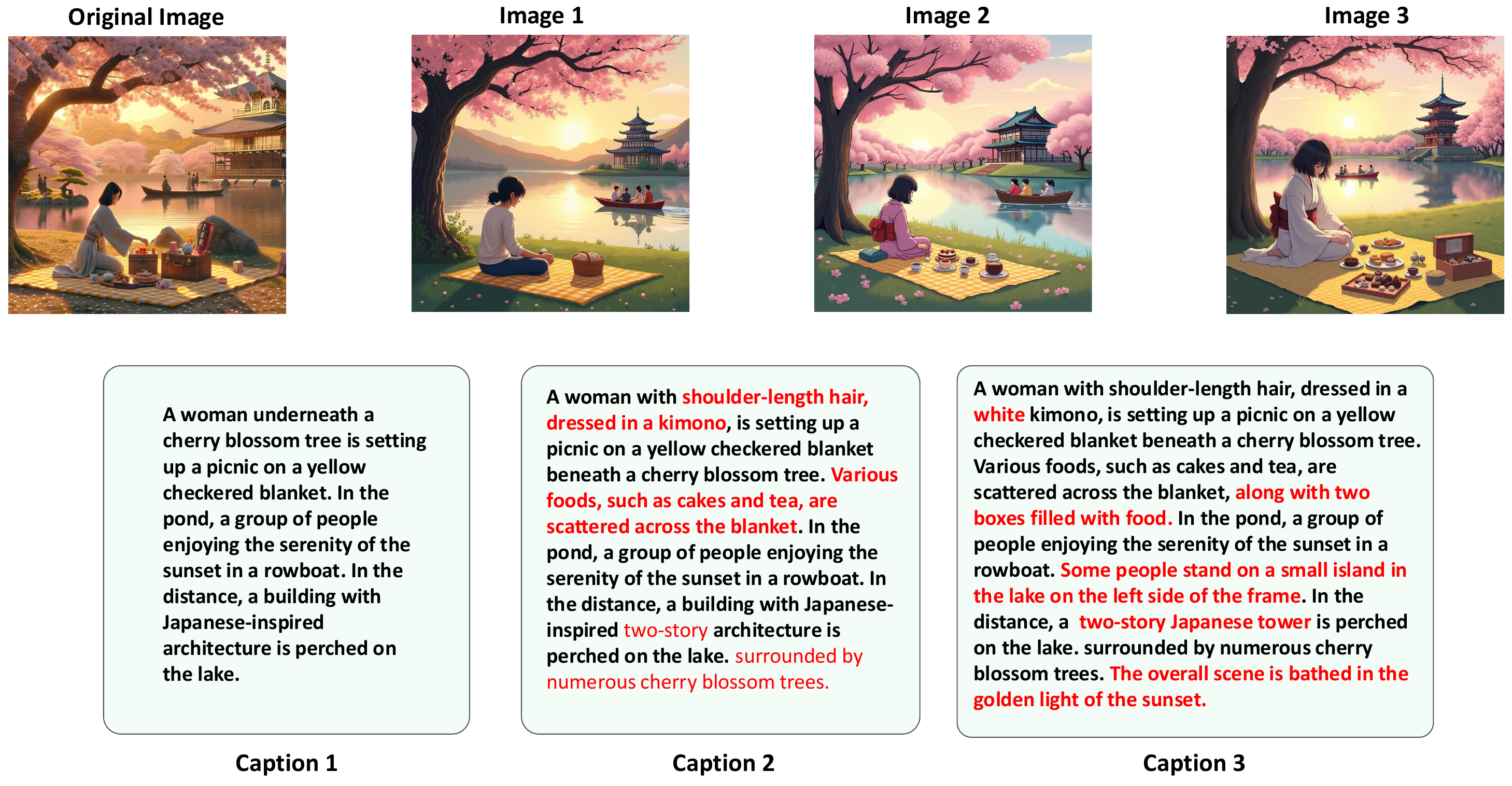}
    \caption{An example demonstrating the iterative refinement process performed by our model, where red text indicates added or corrected information.
}
    \label{fig:app_iter}
\end{figure*}

\begin{figure*}
    \centering    \includegraphics[width=1\linewidth]{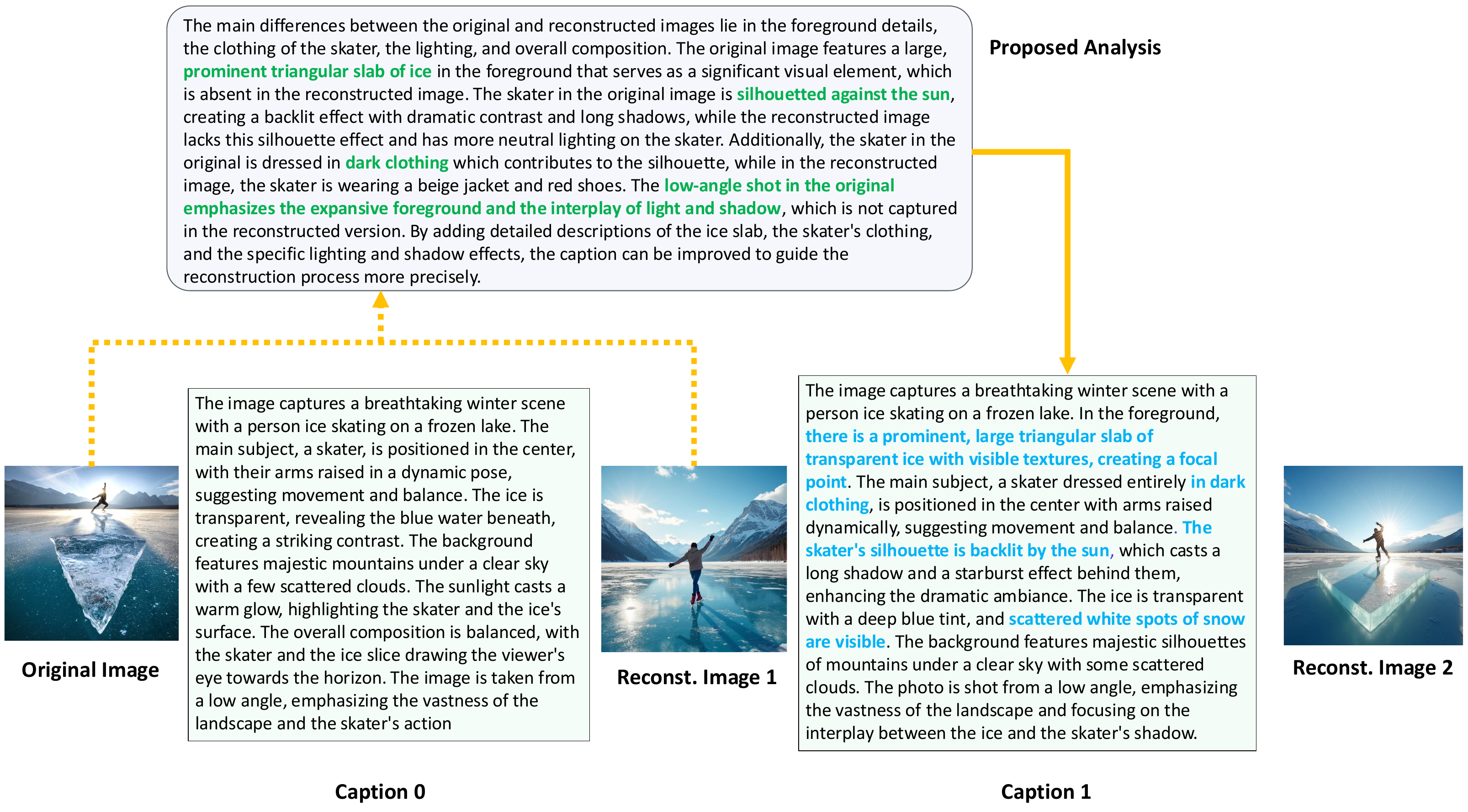}
    \caption{An example demonstrating the output analysis produced by our model, where green text highlights important aspects identified in the analysis, and blue text indicates information incorporated into the updated caption.
}
    \label{fig:app_ana}
\end{figure*}

We present an example of the \modelname refinement process in Fig.~\ref{fig:app_iter}. We can see that as the refinement progresses, the caption is progressively revised to incorporate important missing details. Additionally, Fig.~\ref{fig:app_ana} provides a case accompanied by an in-depth analysis. The analysis illustrates that our refinement model effectively identifies discrepancies and generates reasonable revision suggestions, resulting in more accurate and comprehensive captions.

\subsection{Detailed Results on the Generalization of \modelname}

As discussed in \textsection~\ref{sec:exp_general}, our method consistently performs well across various initial captioning models and prompt configurations. Extended results for different prompt variants are presented in Tab.~\ref{tab:app_prompt}, with the corresponding prompt templates listed in Tab.~\ref{tab:initial_prompt}. Detailed results using different initial captioning models are provided in Tab.~\ref{tab:app_init}. These findings further validate the robustness and effectiveness of \modelname under diverse settings.

\subsection{Detailed Results of the Text-to-Image Generation Experiment}

We present the expanded results of Tab.~\ref{tab: t2i} in Tab.~\ref{tab:t2i_exp}. The text-to-image model trained with captions generated by our method consistently outperforms the model trained with initial captions across nearly all metrics, demonstrating improved alignment between image and text semantic spaces in \modelname.

\section{Additional Information on Experimental Settings}

\subsection{Details of Baselines and Our Implementations} \label{sec:app_base}
We compare our method with several recaptioning baselines. The details of each are provided below:
\begin{table}[ht!]
\caption{Instructions provided to human annotators in the caption editing experiment.
}

\begin{tcolorbox}

{\normalsize \textbf{==  ~\textsc{Instruction to annotators}~ ==}}

\tcblower

We are working on an image captioning task. The following caption was generated by an AI model. Please help refine this caption by correcting any errors or ambiguities based on the image, and feel free to add any important details that are missing from the original caption.

\end{tcolorbox}

\label{tab:human_ins}
\end{table}

\noindent\textbf{LaCLIP}~\cite{laclip} 
~LaCLIP identifies that in CLIP training, text inputs tend to be underutilized due to a lack of augmentation. To address this, the authors propose leveraging large language models (LLMs) to rewrite the given text. Specifically, ChatGPT is used to generate meta input-output pairs, which are then used as in-context examples to prompt LLaMA~\cite{touvron2023llamaopenefficientfoundation} for generating refined captions. In our implementation, we follow the same procedure to obtain enhanced captions. Specifically, we first employ Qwen2-VL-7B-Instruct to simulate the generation of alt text using the prompt: ``Describe the image using a few essential keywords. Keep it concise, within 10 words.'' The meta input-output pairs generated by ChatGPT are then used as in-context examples to prompt Qwen2-VL-7B-Instruct, which generates the final refined captions.

\noindent\textbf{VeCLIP}~\cite{veclip} 
~While previous methods like LaCLIP focus solely on textual rewriting, VeCLIP emphasizes the incorporation of visual concepts into the caption. It first employs a multimodal LLM (LLaVA) to generate captions independently of the original alt text, and then fuses these captions with the original using another LLM, such as Vicuna~\cite{vicuna2023}. In our implementation, we follow the official pipeline. We adopt the same approach as in LaCLIP to generate the initial alt texts. We then utilize LLaVA-1.5-7B-Chat to generate supplementary captions. Finally, Qwen2-VL-7B-Instruct is prompted to fuse these two captions.

\noindent\textbf{CapsFusion~}~\cite{capsfusion} 
~CapsFusion highlights the importance of combining web-based alt texts and synthetic captions. The authors construct a dataset of 1 million examples by prompting ChatGPT to fuse these two types of captions, which is then used to fine-tune LLaMA, resulting in the CapsFusion-LLaMA model. Technically, we adopt the official implementation: we use the same approach as LaCLIP to generate alt texts, utilize Qwen2-VL-7B-Instruct to produce synthetic captions, and apply the official CapsFusion-LLaMA model weights for fusion.

\begin{table*}[]
\small
\caption{Detailed performance of \modelname across different initial captioning models.}
\label{tab:app_init}
\tabcolsep=0.03cm
\begin{tabular}{l|llll|llll}
\toprule[1.5pt]
\multirow{2}{*}{\textbf{Model}} & \multicolumn{4}{c|}{CapsBench}    & \multicolumn{4}{c}{CompreCap}      \\ \cmidrule{2-9}

 & Acc. & Color & Shape & Rel. Pos. & Obj.    & Pix.   & Rel.    & Attr.  \\ \midrule[1pt]

Qwen2-VL (\citeauthor{qwen2vl}) & 42.02 & 48.11 & 27.27 & 32.41 & 69.82  & 60.02  & 2.66 & 2.81 \\
Qwen2-VL + \modelname & 59.02$_\impro{17.00}$ & 67.14$_\impro{19.03}$ & 53.68$_\impro{26.41}$ & 59.51$_\impro{27.10}$ & 75.04$_\impro{5.22}$ & 63.04$_\impro{3.02}$ & 2.85$_\impro{0.19}$ & 2.82$_\impro{0.01}$ \\
\midrule

CogVLM 2 (\citeauthor{hong2024cogvlm2visuallanguagemodels}) & 45.10 & 47.77 & 28.23 & 39.81 & 68.54 & 59.21 & 2.57 & 2.61 \\
CogVLM 2 + \modelname & 57.51$_\impro{12.41}$ & 63.67$_\impro{15.90}$ & 35.46$_\impro{7.23}$ & 48.76$_\impro{8.95}$ & 75.37$_\impro{6.83}$ & 61.65$_\impro{2.44}$ & 2.78$_\impro{0.21}$ & 2.75$_\impro{0.14}$ \\
\midrule

GPT-4o (\citeauthor{openai2024gpt4ocard}) & 49.63 & 54.64 & 28.28 & 48.15 & 70.93 & 60.09 & 2.67 & 2.77 \\
GPT-4o + \modelname & 57.68$_\impro{8.05}$ & 63.24$_\impro{8.60}$ & 44.57$_\impro{16.29}$ & 59.47$_\impro{11.32}$ & 74.47$_\impro{3.54}$ & 62.11$_\impro{2.02}$ & 2.76$_\impro{0.09}$ & 2.81$_\impro{0.04}$ \\
\midrule

Gemini 1.5 Pro (\citeauthor{geminiteam2024gemini15unlockingmultimodal}) & 49.71 & 51.20 & 23.23 & 36.57 & 71.77 & 60.28 & 2.89 & 2.71 \\
Gemini 1.5 Pro + \modelname & 57.72$_\impro{8.01}$ & 65.70$_\impro{14.50}$ & 37.50$_\impro{14.27}$ & 50.48$_\impro{13.91}$ & 75.77$_\impro{4.00}$ & 61.97$_\impro{1.69}$ & 2.85$_\impro{-0.04}$ & 2.83$_\impro{0.12}$ \\
\midrule

BLIP-3~(\citeauthor{xue2024xgenmmblip3familyopen}) & 37.03 & 40.55 & 19.19 & 29.63 & 67.85 & 56.99 & 2.61 & 2.50 \\
BLIP-3 + \modelname & 56.21$_\impro{19.18}$ & 66.20$_\impro{25.65}$ & 37.76$_\impro{18.57}$ & 55.61$_\impro{25.98}$ & 74.31$_\impro{6.46}$ & 61.47$_\impro{4.48}$ & 2.79$_\impro{0.18}$ & 2.75$_\impro{0.25}$ \\
\midrule

LLaVA 1.5 (\citeauthor{llava}) & 29.51 & 27.84 & 9.09 & 18.06 & 57.14 & 44.48 & 2.02 & 2.38 \\
LLaVA 1.5 + \modelname & 53.13$_\impro{23.62}$ & 61.07$_\impro{33.23}$ & 36.84$_\impro{27.75}$ & 48.10$_\impro{30.04}$ & 76.38$_\impro{19.24}$ & 61.49$_\impro{17.01}$ & 2.82$_\impro{0.80}$ & 2.82$_\impro{0.44}$ \\

\bottomrule[1.5pt]
\end{tabular}
\end{table*}

\begin{table*}[]
\caption{Detailed performance of \modelname across different initial prompts.}
\small
\tabcolsep=0.18cm
\label{tab:app_prompt}
\begin{tabular}{l|llll|llll}
\toprule[1.5pt]
\multirow{2}{*}{\textbf{Model}} & \multicolumn{4}{c|}{CapsBench}    & \multicolumn{4}{c}{CompreCap}    \\ \cmidrule{2-9}
     & Acc. & Color & Shape & Rel. Pos. & Obj.    & Pix.   & Rel.    & Attr.   \\ \midrule[1pt]
     
Prompt \#1 & 42.02 & 48.11 & 27.27 & 32.41 & 69.82  & 60.02  & 2.66 & 2.81\\
+ \modelname & 59.02$_\impro{17.00}$ & 67.14$_\impro{19.03}$ & 53.68$_\impro{26.41}$ & 59.51$_\impro{27.10}$ & 75.04$_\impro{5.22}$ & 63.04$_\impro{3.02}$ & 2.85$_\impro{0.19}$ & 2.82$_\impro{0.01}$\\ \midrule

Prompt \#2 & 45.97 & 49.14 & 23.22 & 40.74 & 69.29 & 60.41 & 2.69 & 2.62\\
+ \modelname & 57.64$_\impro{11.67}$ & 65.45$_\impro{16.31}$ & 39.08$_\impro{15.86}$ & 56.45$_\impro{15.71}$ & 74.83$_\impro{5.54}$ & 62.65$_\impro{2.24}$ & 2.80$_\impro{0.11}$ & 2.79$_\impro{0.17}$\\ \midrule

Prompt \#3 & 41.85 & 43.30 & 23.23 & 36.11 & 68.46 & 58.89 & 2.72 & 2.59\\
+ \modelname & 54.85$_\impro{13.00}$ & 66.54$_\impro{23.24}$ & 47.25$_\impro{24.02}$ & 52.85$_\impro{16.74}$ & 75.15$_\impro{6.69}$ & 62.40$_\impro{3.51}$ & 2.80$_\impro{0.08}$ & 2.82$_\impro{0.23}$\\

\bottomrule[1.5pt]
\end{tabular}
\end{table*}

\begin{table*}[ht!]
\caption{Different prompts used to generate initial captions.
}

\begin{tcolorbox}

{\normalsize \textbf{============  ~\textsc{Different prompts to generate initial captions}~ ============}}

\tcblower

Prompt \#1: Describe this image in detail. Your answer should be concise and informative.
~\\

Prompt \#2: Describe the image with rich and detailed observations. You may pay attention to the dimensions of overall, main subject, background, movement of main subject, style, camera movement and so on.
~\\

Prompt \#3: Give this image a detailed caption.

\end{tcolorbox}

\label{tab:initial_prompt}
\end{table*}
\begin{table*}[t] 
\vspace{-0mm}
\label{tab:app_t2i}
\begin{center}
\small
\caption{Extended version of the evaluation of the text-to-image model.
}
	\label{tab:t2i_exp}
	\begin{tabular}{l|ccccc|c}
		\toprule[1.5pt]
        \multirow{2}{*}{\textbf{Model}} & \multicolumn{5}{c|}{DPG-Bench}  &  \multirow{2}{*}{VQAScore}  \\   
        \cmidrule{2-6} 
        & Entity & Relation & Attribute & Global & Overall & \\
        \midrule

        FLUX w/ Init. Cap. & 85.110	& 89.950 &	80.080 &72.414 & 78.502 & 0.841\\
        FLUX w/ \modelname-DPO & \textbf{86.850}	& \textbf{90.551} &	\textbf{82.831}	&\textbf{75.172} & 	\textbf{80.336} & \textbf{0.852}\\

		\bottomrule[1.5pt]
	\end{tabular}
\end{center}

\end{table*}

\noindent\textbf{Self-Loop}~\cite{capture} 
~In the CAPTURE~\cite{capture} paper, the authors introduce a new metric to evaluate image captioning and design a self-looping caption improvement pipeline guided by this metric. In detail, the method detects objects in the image, generates local captions, filters out hallucinated objects, and merges local descriptions with the overall caption. We use the official repository to run this baseline.

\noindent\textbf{ShareGPT4V}~\cite{sharegpt4v} 
~ShareGPT4V underscores the critical role of captions in MLLM training. It uses carefully crafted prompts to guide GPT-4V in generating high-quality descriptions, and then trains a Share-Captioner model to replicate this behavior. In our experiments, we use Share-Captioner to generate captions as part of the baseline comparison.

\subsection{Details of Evaluation Benchmarks}\label{sec:app_bench}
Traditional caption evaluation metrics~\cite{anderson2016spicesemanticpropositionalimage, lin-2004-rouge} are not well-suited for evaluating captions generated by modern MLLMs. In our work, we adopt the following evaluation metrics:

\noindent\textbf{CapsBench.} 
~Proposed in Playground v3~\cite{capsbench}, CapsBench introduces a benchmark designed to evaluate the comprehensiveness and accuracy of image captions. For each image, a set of ``yes-no'' question-answer pairs is generated across 17 semantic categories. During evaluation, an LLM is tasked with answering these questions based solely on the candidate caption. The possible answers are ``yes'', ``no'', and ``n/a'' (for unanswerable questions). The predicted answers are compared with the ground-truth to compute the overall accuracy. This benchmark effectively assesses whether a model can capture accurate and comprehensive information from the image. In our implementation, we use GPT-4o (2024-08-06) as the judge model.

\noindent\textbf{CompreCap.} 
~CompreCap~\cite{comprecap} is a benchmark that evaluates the compositional understanding of detailed visual scenes through a directed scene graph framework. Each image is segmented into semantically meaningful regions, and objects within these regions are annotated with attributes and directional relations to form a directed scene graph. The benchmark then assesses generated captions based on three levels: (1) object-level coverage, (2) accuracy of attribute descriptions, and (3) correctness of key relationships. This benchmark is particularly effective at evaluating the model's ability to capture relational and compositional details. We adopt the official implementation for our evaluation.

\noindent\textbf{Amber.} 
~Amber~\cite{amber} is designed to evaluate hallucinations in MLLM-generated captions by comparing the set of objects mentioned in the caption with a pre-annotated object list for the image. It defines several metrics: \textit{CHAIR}, which quantifies the frequency of hallucinated (i.e., nonexistent) objects, and \textit{Cover}, which measures how well the caption covers the annotated objects. Following the original paper’s claim that ``an ideal response is considered to be one that minimizes hallucinatory content without significantly compromising the coverage of objects in the image,'' we adopt a unified metric, $Cover - CHAIR$, to reflect this trade-off. This provides a concise and interpretable measure of caption faithfulness.

\noindent\textbf{CAPTURE}
~CAPTURE~\cite{capture} introduce a benchmark designed to evaluate detailed image captioning performance by extracting and comparing core visual elements in generated captions. Unlike traditional metrics that rely on n-gram overlaps, CAPTURE focuses on the alignment of semantic content by parsing captions into structured scene graphs comprising objects, attributes, and relationships. Using the Factual T5-based parser~\cite{li2023factualbenchmarkfaithfulconsistent}, these elements are extracted and then matched across candidate and reference captions through a three-stage strategy involving exact, synonym, and soft matching. The final evaluation score is computed as a weighted sum of F1 scores for each element type. This benchmark is particularly effective for assessing fine-grained visual grounding in generated captions. We adopt the official implementation for our evaluation.

\subsection{Details of Recaptioning Experiment with GPT-4o and Human Annotators} \label{sec:app_edit}
In \textsection~\ref{sec:exp_eff}, we analyze the recaption behavior for captions generated by an MLLM. Specifically, we randomly select 100 images from CapsBench and prompt Qwen2-VL-7B-Instruct to generate initial captions. These captions, along with the corresponding images, are then input to GPT-4o, which is asked to refine the captions. For human recaptioning, we recruit two well-educated researchers proficient in English and instruct them to refine the captions. We verify that their edits are consistent and of high quality. The instruction provided to both GPT-4o and the human annotators is shown in Tab.~\ref{tab:human_ins}.
For our \modelname, to ensure a fair comparison, we set the number of refinement iterations $N=1$, i.e., each caption is refined only once.

\subsection{Details of Text-to-Image Generation} \label{sec:app_t2i}
For the text-to-image generation experiment described in \textsection~\ref{sec:exp_t2i}, we adopt the FLUX.1-dev model~\cite{flux2024}. To accelerate training, we employ a LoRA-tuned~\cite{hu2021loralowrankadaptationlarge} version of the model. The training dataset is primarily sourced from Hugging Face~\footnote{\url{https://huggingface.co/datasets/jackyhate/text-to-image-2M}}, and we collect a total of 30K images for our experiments.
Training is conducted for 10{,}000 steps using 8 GPUs, each with a batch size of 10. The image resolution is set to 1024 $\times$ 1024. We use the AdamW optimizer~\cite{loshchilov2019decoupledweightdecayregularization} with a learning rate of $10^{-4}$.

\section{More Implementation Details} \label{sec:app_details}

\subsection{Prompt in the Refinement Procedure} \label{sec:app_prompt}

We provide the prompt used to query GPT-4o~\cite{openai2024gpt4ocard} for the refinement procedure described in \textsection~\ref{sec:method_con}, as shown in Tab.~\ref{tab:gpt4_prompt}.

\begin{table*}[ht!]
\begin{tcolorbox}[breakable]

{\normalsize \textbf{================  ~\textsc{Prompt in the Refinement Procedure}~ ================}}

\tcblower

We are working on a project that involves generating captions for images and using these captions to reconstruct the images. The process follows these steps:

\hangindent=2em\hangafter=1
\textbf{1. Original Image (First Image):} A caption is generated based on this image.

\hangindent=2em\hangafter=1
\textbf{2. Reconstructed Image (Second Image):} The generated caption is used as input for a text-to-image model to create this image.

\bigskip

\textbf{Your Task}

Compare the \textbf{original} and \textbf{reconstructed} images, analyzing their differences to identify potential improvements for the original caption. Based on your observations, provide a \textbf{revised caption} that could enhance the reconstruction quality.

\bigskip

\textbf{Guidelines for Comparison}

\hangindent=2em\hangafter=1
\textbf{• Visual Details:} Color, shape, texture, and material of objects.

\hangindent=2em\hangafter=1
\textbf{• Composition \& Layout:} Object positioning, spatial relationships, and overall scene structure.

\hangindent=2em\hangafter=1
\textbf{• Human Attributes (if applicable):} Pose, facial expression, skin tone, clothing, and hairstyle.

\hangindent=2em\hangafter=1
\textbf{• Perspective \& Style:} Type of image, camera angle, depth of field, lighting, and artistic style.

\hangindent=2em\hangafter=1
\textbf{• Text in the Image:} Accuracy of any visible words, symbols, or signs.

\hangindent=2em\hangafter=1
\textbf{• Image Quality:} Blurriness, artifacts, or inconsistencies in object rendering.

\hangindent=2em\hangafter=1
\textbf{• World Knowledge:} Proper nouns or specific real-world references that should be preserved.

\hangindent=2em\hangafter=1
\textbf{• Color Aesthetics:} Color palette, grading, and overall mood consistency.

\bigskip

\textbf{How to Improve the Caption}

\hangindent=2em\hangafter=1
\textbf{• Add missing details} that were lost in reconstruction.

\hangindent=2em\hangafter=1
\textbf{• Clarify ambiguous descriptions} to provide more precise information.

\hangindent=2em\hangafter=1
\textbf{• Correct any inaccuracies} based on observed differences.

\hangindent=2em\hangafter=1
\textbf{• Specify key attributes} (e.g., “a red leather couch” instead of “a couch”).

\bigskip

Your revised caption should aim to \textbf{reduce discrepancies} between the original and reconstructed images while maintaining a natural and informative description. You are encouraged to make the revised caption less than 512 tokens.

\bigskip

Now I provide the original image, reconstructed image, and the original caption: \texttt{\{orig\_caption\}}.

\bigskip

Please give me the revised caption that you believe could enhance the reconstruction quality (i.e., make the new reconstructed image more like the original one at pixel level), enclosed with \texttt{<revised caption>}. And provide your analysis enclosed with \texttt{<analysis>} after.

\end{tcolorbox}
\caption{The prompt used to query GPT-4o in the refinement procedure.}
\label{tab:gpt4_prompt}
\end{table*}

\subsection{Details of DPO Training}

For training the DPO variant, we primarily use data from the DCE dataset~\cite{sun2025descriptivecaptionenhancementvisual}, which spans a diverse range of image domains. From this dataset, we randomly sample 10K instances to construct preference pairs, as outlined in \textsection~\ref{sec:method_dpo}.

For the DPO experiments, we use the \texttt{LLaMA-Factory} toolkit~\cite{zheng2024llamafactory}. We initialize the model with Qwen2-VL and set the preference scaling parameter to $\beta = 0.1$. The model is fine-tuned for 3 epochs using 8 GPUs. The batch size is set to 64, and the learning rate is $\eta = 1.0 \times 10^{-5}$. We use a cutoff length of 2048 tokens and a warmup ratio of 0.1.

\section{Basics for DPO}

Direct Preference Optimization (DPO)~\cite{rafailov2024direct} formulates preference learning as a probabilistic binary classification task, without the need to train an explicit reward model. Given a dataset of preference tuples $(x, y^+, y^-)$---where $x$ denotes a shared context (e.g., a prompt), and $y^+$ and $y^-$ represent the preferred and dispreferred responses respectively--DPO aims to train a policy $\pi_\theta(y \mid x)$ such that:
\[
\pi_\theta(y^+ \mid x) > \pi_\theta(y^- \mid x)
\]

DPO defines an implicit reward function based on the log-likelihood ratio between the current policy $\pi_\theta$ and a fixed reference policy $\pi_0$ (e.g., the base model):
\[
r(y \mid x) = \log \frac{\pi_\theta(y \mid x)}{\pi_0(y \mid x)}
\]

This leads to a binary classification objective that maximizes preference likelihood with KL regularization:
\[
\begin{aligned}
\pi^* &= \arg\max_{\pi}  \mathbb{E}_{(x, y^+, y^-)} \\ &\Bigg[ 
\log \frac{\exp(\beta r(y^+ \mid x))}{\exp(\beta r(y^+ \mid x)) + \exp(\beta r(y^- \mid x))} 
\Bigg] \\
& - \mathrm{KL}(\pi \,\|\, \pi_0)
\end{aligned}
\]

Substituting $r(y \mid x)$, the DPO training loss becomes:
\[
\begin{aligned}
\mathcal{L}_{\mathrm{DPO}} &= - \mathbb{E}_{(x, y^+, y^-)} \\&\Bigg[ 
\log \frac{\pi_\theta(y^+ \mid x)^\beta}
{\pi_\theta(y^+ \mid x)^\beta + \pi_\theta(y^- \mid x)^\beta}
\Bigg]
\end{aligned}
\]

This loss encourages the model to prefer $y^+$ over $y^-$ while implicitly regularizing against the reference model $\pi_0$. Unlike traditional reinforcement learning, DPO requires no reward model sampling or rollouts, offering both stability and efficiency. More mathematical details can be found in the original paper of DPO.\footnote{\url{https://arxiv.org/pdf/2305.18290}}

\section{Limitations}

Our work still has several limitations. First, a key assumption of the proposed pipeline is that the text-to-image model must be sufficiently powerful to faithfully recover as many details as possible from the candidate caption. This places high demands on the capability of the text-to-image model. In this work, we adopt the FLUX model, which demonstrates strong performance, but still leaves room for significant improvement. Secondly, given the discrepancies between the original and reconstructed images, multiple plausible caption revisions may exist. Determining how to refine the caption in a concise yet effective manner remains a significant challenge for the refinement model. Lastly, the iterative version of our method is resource-intensive. Although we propose a DPO-based variant to mitigate this issue, reducing the coupling within the pipeline and improving inference efficiency remain important directions for future work.
\end{document}